\documentclass[10pt,twocolumn,letterpaper]{article}

\usepackage{iccv}
\usepackage{times}
\usepackage{epsfig}
\usepackage{graphicx}
\usepackage{amsmath}
\usepackage{amssymb}
\usepackage{stmaryrd}

\usepackage[table,x11names]{xcolor}
\usepackage{color}
\usepackage{colortbl}
\usepackage[accsupp]{axessibility}

\definecolor{LightCyan}{rgb}{0.88,1,1}
\definecolor{LightGray}{rgb}{0.95,0.95,0.95}

\usepackage[pagebackref=true,breaklinks=true,letterpaper=true,colorlinks,bookmarks=false]{hyperref}

\iccvfinalcopy 

\def\iccvPaperID{10587} 

\ificcvfinal\fi

\newif\ifdraft
\drafttrue

\newcommand{\myparagraph}[1]{\smallskip\noindent\textbf{#1}}

\begin{document}

\title{
Multi-Class Cell Detection Using Spatial Context Representation
}

\author{Shahira Abousamra, David Belinsky, John Van Arnam, Felicia Allard, Eric Yee, 
\\
Rajarsi Gupta, Tahsin Kurc, Dimitris Samaras, Joel Saltz, Chao Chen 
\\
Stony Brook University\\ 
Stony Brook, NY 11794, USA
}


\maketitle
\ificcvfinal\thispagestyle{empty}\fi

\begin{abstract}
In digital pathology, both detection and classification of cells are important for automatic diagnostic and prognostic tasks. Classifying cells into subtypes, such as tumor cells, lymphocytes or stromal cells is particularly challenging. Existing methods focus on morphological appearance of individual cells, whereas in practice pathologists often infer cell classes through their spatial context. In this paper, we propose a novel method for both detection and classification that explicitly incorporates spatial contextual information. We use the spatial statistical function to describe local density in both a multi-class and a multi-scale manner. Through representation learning and deep clustering techniques, we learn advanced cell representation with both appearance and spatial context. On various benchmarks, our method achieves better performance than state-of-the-arts, especially on the classification task. We also create a new dataset for multi-class cell detection and classification in breast cancer and we make both our code and data publicly available.
\end{abstract}

\section{Introduction}
\label{sec:introduction}
We propose the first joint cell detection and classification method that explicitly learns a spatial-context-aware representation of cells. We demonstrate that incorporating spatial context will significantly improve the performance, especially for the classification task.

Identification of various types of cells such as tumor cells, lymphocytes, and stromal cells from whole-slide histology images is an important step towards automatic diagnosis and prognosis in digital 
pathology. 
The spatial arrangement of different cells can comprehensively characterize the interaction between tumor and immune cells and be correlated with clinical outcomes \cite{nawaz2016computational,Yuan:2012:cell-classify,Kavianpour:2020:classification-importance}. One good example is the detection and measurement of tumor infiltrating lymphocytes (TILs), i.e., lymphocytes residing within the border of invasive tumors \cite{salgado2015evaluation}. The prevalence of TILs has been shown to be associated with better clinical outcomes \cite{saltz2018spatial,Shibutani:2018:imp-classify:TIL-outcome,Stanton:2016:imp-classify:TIL-outcome}. 
Aside from lymphocytes, the presence of isolated or small clusters of tumor cells at the invasive tumor front, a phenomenon known as tumor budding, is a prognosis biomarker associated with an increased risk of lymph node metastasis in colorectal carcinoma and other solid malignancies \cite{lugli2020tumour}. 
Other examples include the assessment of lymphovascular invasion and perineural invasion \cite{li2015prognostic} and the identification and measurement of intraepithelial lymphocytes for the diagnosis of celiac disease \cite{robert2018statement}. 
All these studies necessitate an effective algorithm to accurately identify cells of different types.

Multi-class cell identification involves both cell detection and cell classification.
Cell detection has been studied extensively in the past few decades \cite{sirinukunwattana:itmi:2016:point_class,Hofener:2018:cell-detection,Xu:2016:ITMI:cell-detection}.
Existing approaches either adopt the object detection algorithm from computer vision \cite{Hung:2020:BMC:nucleus-detection:faster-rcnn,Yousefi:2019:ISBI:cell-detect:faster-rcnn}, 
or treat the problem as an instance segmentation problem and segment nuclei one-by-one \cite{Graham:2020:ITMI:nucleus-segmentation,graham:2019:hover,RAZA:2019:MIA:nucleus-segmentation,Kumar:2017:ITMI:nucleus-segmentation,naylor:2018:ipmi:cell-sup-seg}.  
Although segmentation methods provide detailed nuclei morphology, their training requires highly detailed and thus time-consuming nuclei mask annotation.
To circumvent this bottleneck, 
one may use weakly-supervised methods \cite{Qu:midl:2019:point_seg,yoo:miccai:2019:point_seg,Tian:miccai:2020:point_seg,chamanzar:isbi:2020:point_seg} to segment nuclei based only on \emph{point annotations}, i.e., points placed at the centers of nuclei. Point annotation is a much more affordable annotation for large scale training.



Despite the success in the cell detection, our progress on cell classification is not as advanced.
Indeed, classification is a challenging task even for human experts.
Cells of different kinds can manifest with similar appearance. 
Meanwhile, cells of the same type may exhibit large variation of morphology and texture in regions of neoplasia and inflammation. 
To correctly classify cells in such challenging scenarios,  
pathologists not only use appearance, but also rely on the contextual information of surrounding cells, their spatial relationships and tissue architecture.
For example, degenerating or apoptotic cells often cluster together within the luminal spaces of gland forming tumors and can be readily identified in this context despite the spectrum of morphologic features they display; a context which can be identified through cellular architectural patterns in a larger scale.  Similarly, architectural patterns can be used to help distinguish reactive stromal cells from tumor cells when their shape and chromatin pattern are indistinguishable.

To design an ideal classification algorithm, it is essential to imitate a pathologist’s thought process and incorporate the spatial context into the decision making.  
Unlike existing methods which only learns the context implicitly  \cite{graham:2019:hover,sirinukunwattana:itmi:2016:point_class},
we propose a novel algorithm that explicitly leverages spatial context.
To model the spatial context, we introduce the classic Ripley's K-function \cite{dixon2014r} from spatial statistics \cite{Baddeley:2015:book:spatial-point-pattern}. The K-function encodes spatial relationship between cells in a multi-class, multi-scale manner. It has been shown to be a powerful descriptor of cellular architectures  \cite{Yuan:2012:cell-classify,bull2020combining}. However, existing studies use the K-function mainly for downstream analysis, not for better cell identification as we do.

\begin{figure}[t]
    \begin{center}
      \includegraphics[width=1\linewidth]{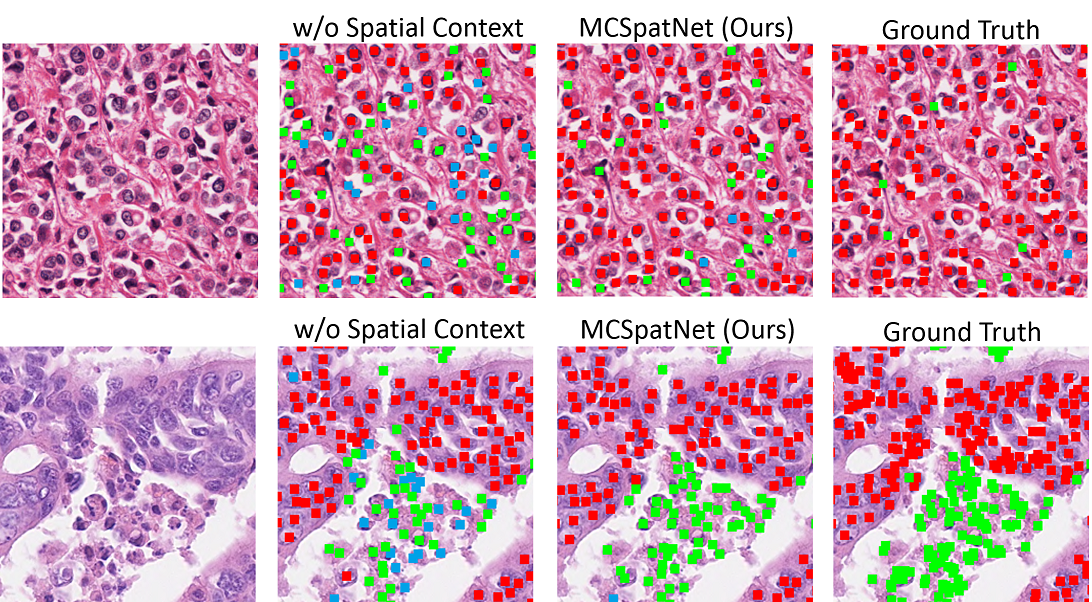}
  \end{center}
  \caption{Sample results with and without spatial contexts. Blue, red, and green dots represent inflammatory, epithelial, and stromal cells, respectively. From left to right: original patches, detection and classification results without spatial contexts, our method using spatial contexts, ground truth point annotation. Our method better classify cells using spatial contexts. }
\label{fig:teaser}
\end{figure}

Our major challenge is that the spatial context is not available at inference time; one cannot infer the K-function without first identifying cells and their classes. To address this challenge, we assume a deep neural network has sufficient learning power and \textbf{propose to learn a spatial-context-aware representation through a multi-task learning framework.} We train a deep neural net that jointly performs cell detection, cell classification and spatial context prediction (i.e., predicting K-functions). Through training, the network learns a representation that incorporates both appearance and spatial context. At inference stage, only the detection and classification modules are used. But the spatial-context-aware representation ensures a superior performance.
See Figure \ref{fig:teaser} for an illustration.

To learn with such a multi-task framework is challenging, because the tasks are very different in nature. The cell classification/detection modules predict class labels. Whereas the spatial context prediction module needs to predict K-functions, which are high-dimensional continuous-valued vectors.
To better learn with these tasks, we introduce a deep clustering module, inspired by works in unsupervised and weakly-supervised learning \cite{caron:2018:eccv:deep-cluster,Chang:2020:CVPR:deep-clustering:weak-supervised}. In this module, 
we introduce pseudo-labels generated by clustering of the deep representation. These pseudo-labels serve as a connection between cell class labels and K-functions, thus facilitate the collaboration of different modules and the fusion of appearance and spatial information.
See Figure \ref{fig:arch-wide} for an overview of our model.

We apply our method, called MCSpatNet, to a joint cell detection and classification task.
We evaluate our method on three benchmark datasets (breast cancer, colorectal cancer and lung cancer), all of which with multi-class point annotations.\footnote{Our model can naturally be extended to mask annotations. In this paper, we focus on point annotations as they are much more affordable.
Indeed, our experiments reveal that our spatial-context-aware model trained with point annotations can be as good as segmentation-based models trained with mask annotations w.r.t.~detection and classification. 
} Our method outperforms various SOTA baselines, demonstrating the power of the spatial-context-aware representation.

To summarize, our contributions are as follows: 
\begin{itemize}
    \item We propose a novel cell detection and classification method which, for the first time, explicitly learns a spatial-context-aware representation of cells via multi-task learning.
    \item We introduce spatial statistical functions (K-function) as an effective descriptor of cells' spatial context.
    \item We utilize spatial context prediction module and deep clustering module to facilitate learning of the representation.
    \item We create a new dataset for multi-class cell detection and classification in H\&E images of breast cancer (BRCA-M2C). The dataset is available here:
    \url{https://github.com/TopoXLab/MCSpatNet}       
    \item Our code is available here: \url{https://github.com/TopoXLab/MCSpatNet}   
\end{itemize}


    

\subsection{Related Work}
\label{sec:related-work}


Most cell detection from point annotation methods operate by training a regression network to predict a probability density map that has high values at the ground truth points and the probability attenuates as you go further away. The detected cells are then the peaks in the regressed maps 
\cite{sirinukunwattana:itmi:2016:point_class,Hofener:2018:cell-detection}. 
In \cite{Xu:2016:ITMI:cell-detection,Khoshdeli:2017:EBS:cell-detect} a patch-wise classification is performed of whether a patch contains a cell nucleus in the center. \cite{Hung:2020:BMC:nucleus-detection:faster-rcnn,Yousefi:2019:ISBI:cell-detect:faster-rcnn} use faster-RCNN object detection algorithm \cite{faster-rcnn:2018:NIPS}. 
One can naturally extend the method by using other similar object detection algorithms \cite{cai:cascade-rcnn:2018:CVPR,cheng:revisit-rcnn:eccv:2018,cai:cascade-rcnn2:tpami:2019}. 
Our proposed method learns to detect cells by pixel-wise binary classification. It is trained with ground truth binary maps where each connected component represents a cell nucleus and the components are restricted to not overlap so as to respect the boundaries between cells. 

When nuclei segmentation masks are available, fully supervised instance segmentation methods can be trained to predict nuclei contours \cite{zhou:2019:ipmi:cia:cell-sup-seg, graham:2019:hover,naylor:2018:ipmi:cell-sup-seg}.
Alternatively, weakly supervised nucleus segmentation methods learn to segment nuclei using ground truth point annotations \cite{Qu:midl:2019:point_seg,yoo:miccai:2019:point_seg,Tian:miccai:2020:point_seg,chamanzar:isbi:2020:point_seg}. These methods mostly implement some form of detection followed by refinement to get more accurate boundaries.

There are fewer methods that target cell classification, mostly operating in 2-stages. \cite{sirinukunwattana:itmi:2016:point_class} has a 2-stage detection and patch-wise classification, where the patches are small windows centered at the detected cells.  \cite{Chang:2017:cell-classifcation,Yuan:2012:cell-classify} perform a 2-stage segmentation and patch-wise classification. \cite{Yuan:2012:cell-classify} further applies spatial kernel smoothing and cancer region detection to improve the classification accuracy. \cite{zormpas2019superpixel} also applies a classification smoothing on top of \cite{sirinukunwattana:itmi:2016:point_class} using conditional random field (CRF) built on cells and superpixels. 
\cite{Zhang:2017:cell-classifcation} classifies readily available patches that are coarsely centered around the nuclei. \cite{graham:2019:hover} proposes a fully supervised segmentation and pixel-wise cell classification method using 3 network branches for segmentation, boundary detection, and classification tasks. 

Deep clustering \cite{caron:2018:eccv:deep-cluster,asano:2020:iclr:deep-cluster,PCL:Li:2021:iclr:deep-cluster,Ji:2019:ICCV:deep-cluster} has been known to enhance feature representation in unsupervised and weakly supervised learning \cite{Ji:2019:ICCV:deep-cluster,wang:2020:neurocomputing:deep-clustering:weak-supervised,Chang:2020:CVPR:deep-clustering:weak-supervised}. Mostly, deep clustering methods generate clusters from feature representations and learn to predict the cluster for each input instance  \cite{caron:2018:eccv:deep-cluster,PCL:Li:2021:iclr:deep-cluster}. Other methods try to learn clusters that maximize the information across classes \cite{asano:2020:iclr:deep-cluster,Ji:2019:ICCV:deep-cluster}. Similar to \cite{caron:2018:eccv:deep-cluster,Chang:2020:CVPR:deep-clustering:weak-supervised}, we iteratively generate pseudo labels for sub-classifying cells by deep clustering. We use features encoding the spatial statistics functions and the class texture features thus combining the spatial and visual contexts for a better representation.

\section{Method}
\label{sec:method}
\begin{figure*}[t]
    \begin{center}
       \includegraphics[width=1\linewidth]{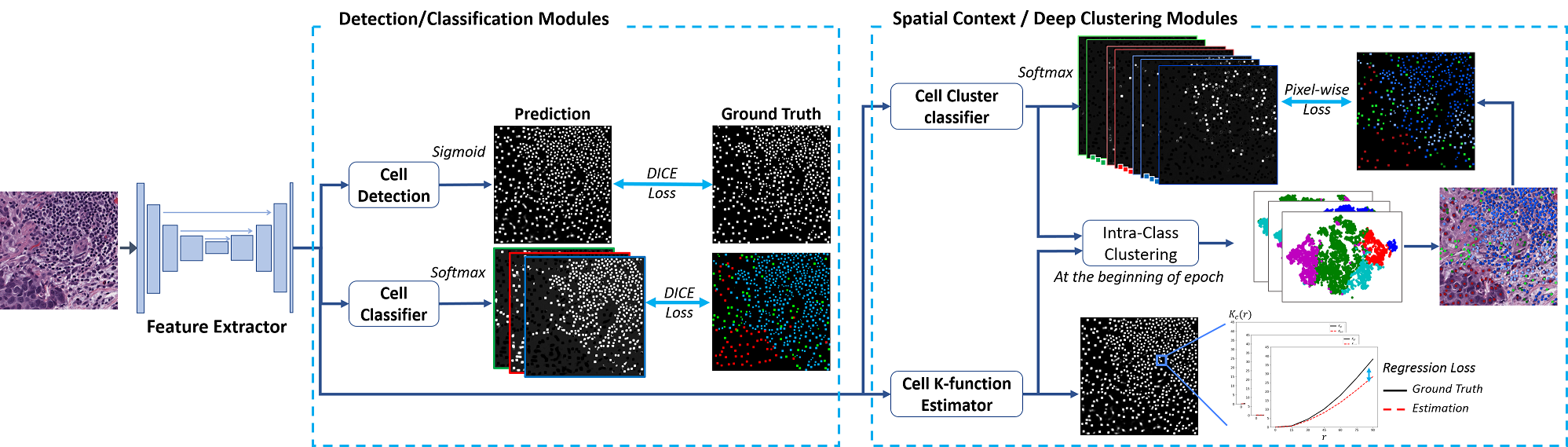}
   \end{center}
   \caption{Model architecture. The feature extractor generates shared features for all four following modules: cell detection, classification, spatial context, and deep clustering. 
   Each module has its own layers to generate its own task-specific representation.
   The spatial context module learns the K-function of each cell. The deep clustering module performs dynamic clustering based on spatial context representations and cell classification representations. The model is trained end-to-end and thus learns a spatial-context-aware feature representation.}
\label{fig:arch-wide}
\end{figure*}

\begin{figure}[t]
    \begin{center}
       \includegraphics[width=\linewidth]{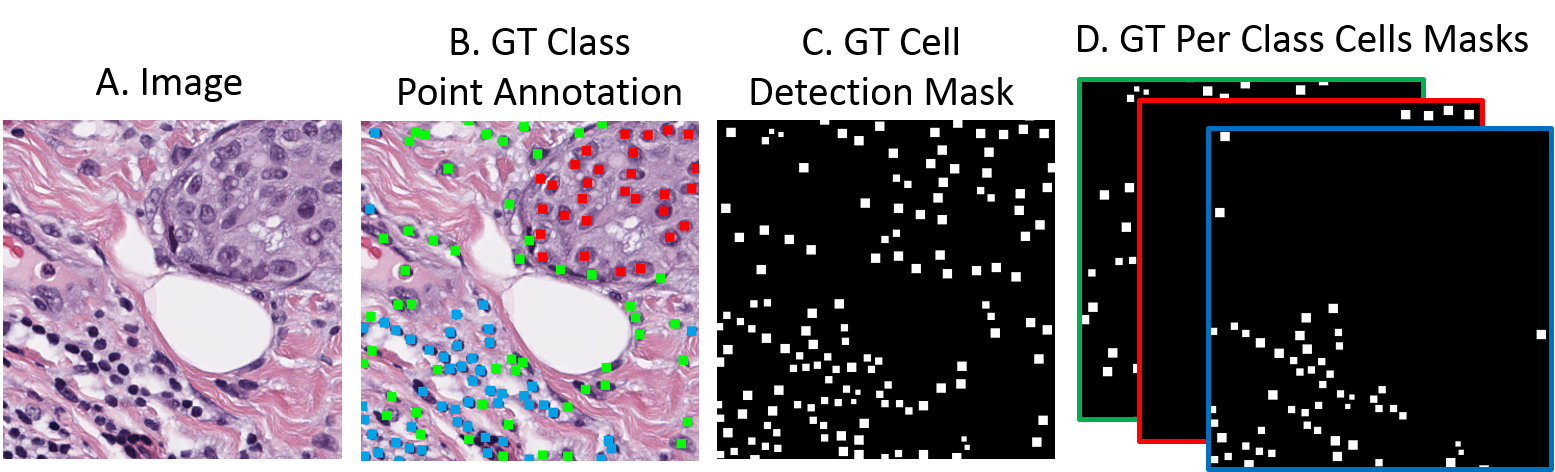}
   \end{center}
   \caption{Input and ground truth (GT) training maps. (A) Sample training patch. (B) The GT point annotation.
   (C) The detection GT mask. (D) The classification GT masks. Blue,
red, and green indicate inflammatory, epithelial, and stromal
cells/channels, respectively. }
\label{fig:gt_masks}
\end{figure}

We propose a method for joint cell detection and classification on H\&E stained images. The prediction and the ground truth are in the form of multi-class point annotations; one point is positioned in the approximate center of each nucleus, with specific cell class labels: inflammatory, epithelial\footnote{Our patches are selected from cancerous regions, in which epithelial cells are tumor cells.}, and stromal cells. See Fig.~\ref{fig:gt_masks} for an illustration. 


Our model has several different modules corresponding to different tasks. These modules share the same input and the same feature extractor. But they have their own blocks of convolutional layers for their own different prediction tasks. This way these tasks learn and benefit from a common representation without conflicts.

The architecture is shown in Fig.~\ref{fig:arch-wide}. Aside from the cell detection and classification modules, our method has two additional modules.
First, we introduce a spatial distribution prediction module, which learns to predict the cells' associated spatial statistics functions. As a result, it learns to pool features that describe the spatial context. 
To further improve the feature representation, we introduce a cell-level deep clustering module. The module iteratively cluster cells based on the feature representation and predict their clusters. It integrates both appearance and spatial contexts into a better feature representation. 

In Sec.~\ref{sec:method:spatial-context}, we formulate the spatial context information, and illustrate its intuition. In Sec.~\ref{sec:method:spatial-learning}, we provide details of all four modules and the common feature extractor. 

\subsection{Cellular spatial context}
\label{sec:method:spatial-context}
\begin{figure}[t]
    \begin{center}
       \includegraphics[width=0.9\linewidth]{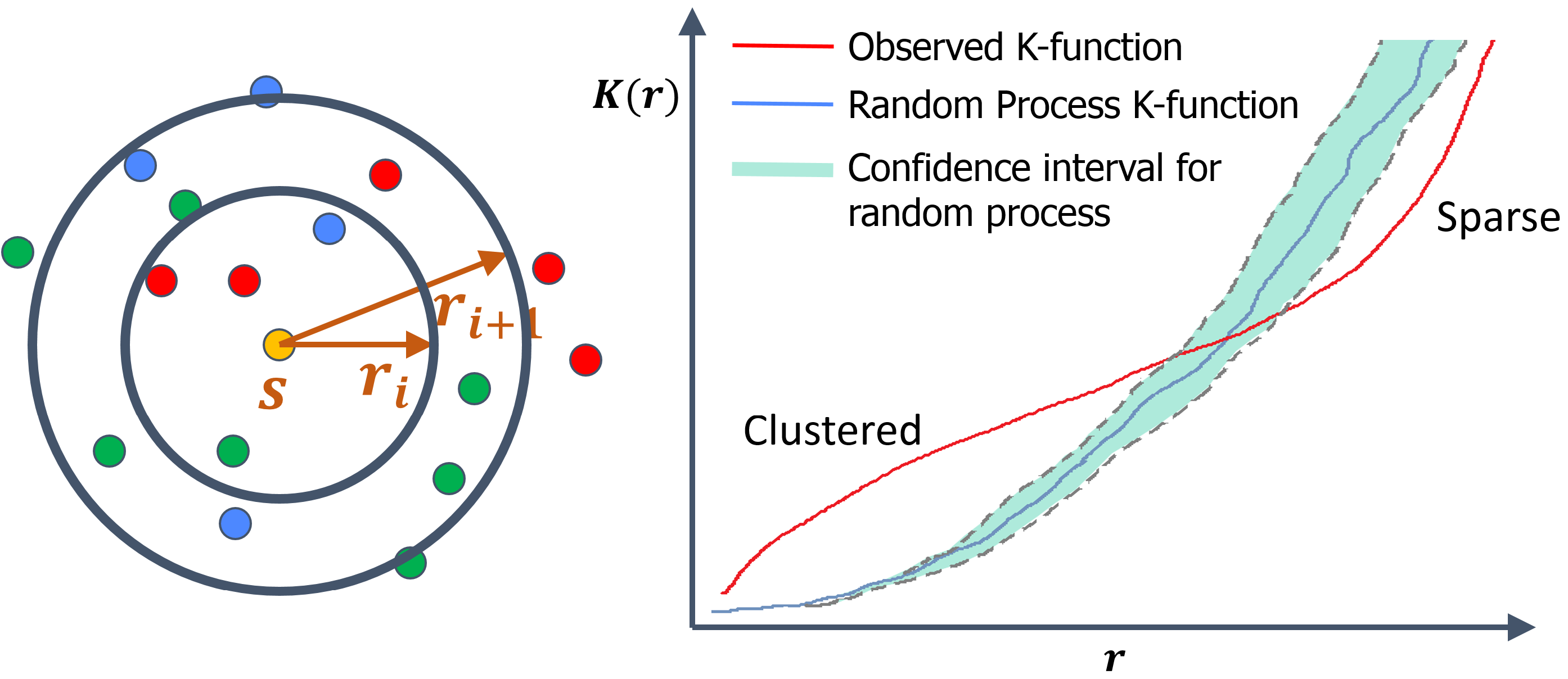}
   \end{center}
   \caption{Ripley's K function. Left: The K-function considers the number of neighboring target points (cells) of different classes at increasing radii from a source (cell) $s$. The sources and targets can each be points from a specific class. Right: Plot of an observed K-function together with the baseline from a Poisson point process. The dashed lines are the confidence interval envelopes for the random process function. 
   }
\label{fig:k-func}
\end{figure}

We define the spatial context of a cell as the distributions of cells of different classes in its surrounding neighborhood. To describe the spatial context, we introduce \textbf{Ripley's K-function}, a spatial statistics function that describes point patterns \cite{dixon2014r,Baddeley:2015:book:spatial-point-pattern}. 
For a cell of interest, called the \emph{source}, we can measure the number of neighbors (called targets) within a distance $r$ from the source. 
Aggregating over all observed points, we have the K-function as the cumulative distribution function that represents the expected number of neighbors within increasing distances $r_i$ from the source. 
See the illustration in Fig.~\ref{fig:k-func}. 
Formally, given a 2D point set $X$ of size $n$, the K-function is: 
\begin{equation}
K(r) = \frac{1}{\lambda} \sum\nolimits_{s\in X} \sum\nolimits_{t\in X\backslash\{s\}}\frac{1}{n-1} \llbracket d(s,t) < r \rrbracket
    \label{eq:k-func}
\end{equation}
where $d(s,t)$ is the Euclidean distance between the source and target points $s$ and $t$. $\llbracket \cdot \rrbracket$ is the Iverson bracket, which is one if the condition inside it is true, and zero otherwise. $\lambda$ is the intensity function used to normalize w.r.t.~the density of the source points. Depending on the assumption, the intensity function can be a constant, i.e., $\frac{Area}{n}$, (homogeneous setting) or location dependent (inhomogeneous).

We can compare the calculated K-function with the baseline, i.e., the K-function of a random point process (Poisson point process). When the K-function of interest is above the baseline at a certain range of $r$, we know there are more target points than one would expect from a random point process, and thus the points are clustered. When the K-function is below the baseline, we have less target points than one would expect from a random point process, hence the points are dispersed. 
For multi-class point sets, we extend the K-function so that the source and target can be from different classes of points. In such case, the K-function is also called the K-cross function. 

\myparagraph{Cell-specific K-functions and vectorization.} 
In cell detection and classification, we focus on individual cells and inspect the spatial context of each cell rather than at the population level. For a given cell $s$ as the source, we restrict to a fixed-size region surrounding it. We only consider target cells falling within this local region or patch. We consider target cells of different classes one-by-one. For class $c$, denote by $X^c_s$ all class-$c$ cells within the local patch centered at $s$. The K-function of class $c$ is
\begin{equation}
K^c_s(r) = \frac{1}{n_{max}}\sum\nolimits_{t\in X^c_s\backslash\{s\} } \llbracket d(s,t) < r \rrbracket
    \label{eq:cell-k-func}
\end{equation}
where $n_{max} = \max_{s}\sum_c |X^c_s| $ denotes the maximum number of target cells for any local patch. 

In practice, we set the patch size to $180\times180$. We also vectorize the K-function by uniform sampling at a finite set of radii: $r = 15, 30, 45, \ldots, 90$ pixels. In total, we have three classes, each with six dimension. 
We call this 18 dimensional vector the \textbf{K-function vector} of $s$. By learning to predict this K-function vector, our model learns the spatial representation.

\begin{figure}[t]
    \begin{center}
       \includegraphics[width=0.8\linewidth]{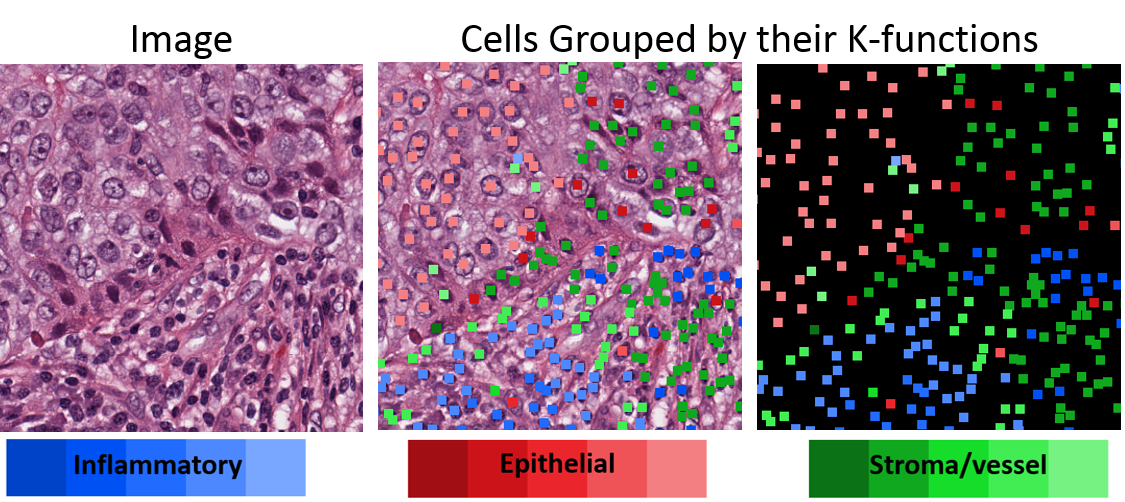}
   \end{center}
   \caption{Visualization of cells with distinct K-functions. We group cells into sub-categories based on their K-function vectors. Different sub-categories exhibit different spatial behavior. Please note that the choice of darker to lighter colors does not imply a sequential relationship between sub-categories. }
\label{fig:k-func-clustering}
\end{figure}

\myparagraph{K-functions and cells' spatial behavior.}
We finalize this subsection by providing real examples to illustrate how K-functions can help refine cells' spatial representation.
In Fig.~\ref{fig:k-func-clustering}, we visualize different classes of cells and their K-functions. Since K-function is high dimensional, we cannot directly show their values. Instead, we cluster cells of each class into sub-categories based on their K-function vectors, and then visualize different sub-categories with different colors (from dark blue to light blue for inflammatory cells, from dark green to light green for stromal cells, from dark red to light red / pink for epithelial cells).

\begin{figure}[t]
\begin{center}
   \includegraphics[width=1\linewidth]{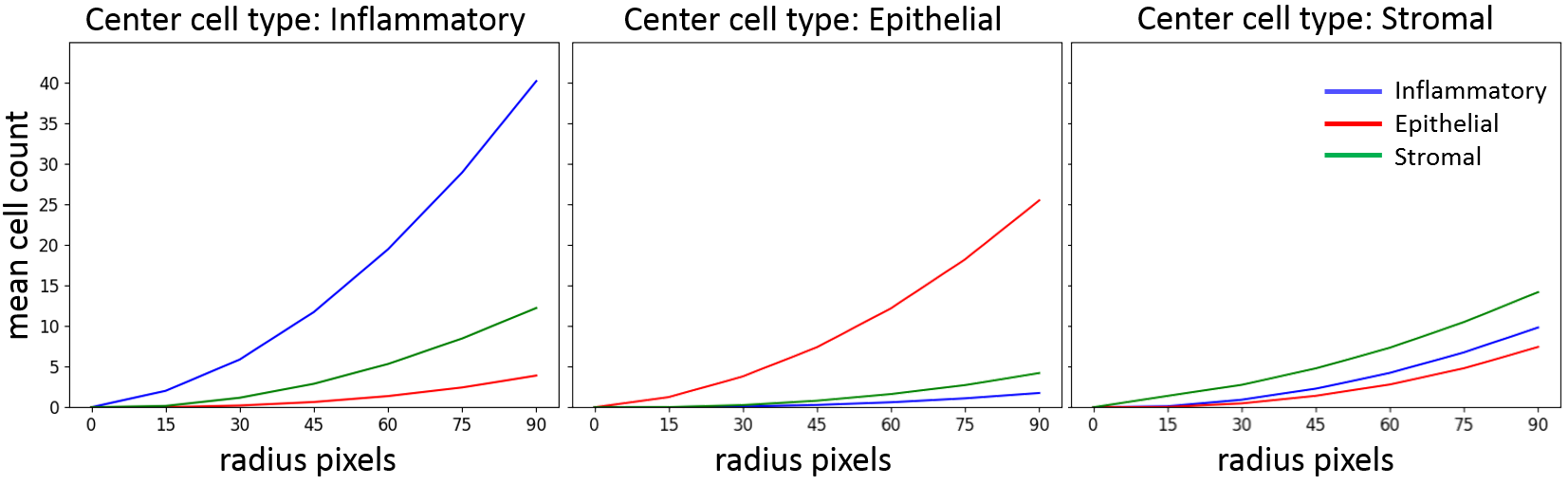}
\end{center}
   \caption{Average K-functions when the centers are inflammatory (left), epithelial (center), and stromal cells (right). In each plot, blue, red, and green curves represent different neighboring cell types.}
\label{fig:k-func-avg}
\end{figure}

We observe that different sub-categories clearly exhibit distinct spatial behavior.
For epithelial cells, cells clustered within a tumor nest often belong to the pink sub-category, whereas those that are more dispersed and closer to stromal cells belong to the red sub-category. 
For stromal cells, the ones from the light green sub-category are often closer to other classes (inflammatory or epithelial cells). Meanwhile, other stromal cells that are far from inflammatory/epithelial cells tend to belong to the dark green sub-category. Similar behavior can be observed for inflammatory cells: light blue cells are clustered while dark blue cells are dispersed.  
Moreover, Fig.~\ref{fig:k-func-avg}, shows the average K-functions for the different pairs of source and target cell classes. It is obvious that different pairs of cells classes exhibit different spatial behavior. 

This demonstrates that K-functions do stratify cells into sub-categories with distinct spatial behavior. It motivates us to learn the spatial representation via K-functions. 


\subsection{Multi-task learning for a spatial-context-aware representation}
\label{sec:method:spatial-learning}
Our proposed model has four modules for four different tasks: cell detection, cell classification, deep clustering, and spatial context prediction.
The four modules share a common feature extractor that learns a shared feature representation. The deep clustering module and the spatial context prediction module are only used in training. They help learn a spatial-context-aware representation which improves the performance of cell detection and classification. Please see Fig.~\ref{fig:arch-wide} for the model architecture.

\myparagraph{Feature extractor.}
The feature extractor is a variant of U-Net model \cite{unet:2015:miccai} with a VGG-16 \cite{vgg:Zisserman:2015:ICLR} backbone. 
The output of the feature extractor has 96 channels and the same spatial resolution as the input. 
The output feature representation is shared by all four task-specific modules.

Each of the task-specific modules also has its own block of convolutional layers.
For all the task-specific blocks, the spatial resolution of the input and output are the same as the input image. But the number of output channels are different for different tasks. 
This allows different task-specific blocks to tune the features to better suit their own tasks without conflict. 
The tasks are very different; involving both classification and regression. The deep clustering task is very dynamic in nature. Without its own task-specific block, it will destabilize the feature extractor and affects other tasks negatively.

Next, we explain different task-specific modules.

\myparagraph{Cell detection and classification modules.}
For cell detection, the model predicts a single channel likelihood map over all pixels and compare it with a binary ground truth mask.  
The ground truth mask is generated by dilating the point annotation slightly. For cells close to each other, we use smaller dilation radii to avoid overlapping. Each connected component in the mask corresponds to one cell. See Fig.~\ref{fig:gt_masks} for an illustration. 
The output of the cell-detection block has a single channel with sigmoid activation.
The training for this module uses DICE loss as it tends to preserve small objects in the image.

For the classification task, we create a similar ground truth mask as the detection ground truth mask, except that pixels in each connected component has a specific cell class. The cell-classification block output has three channels, corresponding to the three cell classes. We use a softmax activation and use DICE loss for training.

During inference, we only use the detection and classification branches. We threshold the cell-detection block output and use the centroids of all connected components as the predicted cell locations. We classify each cell using the cell-classification block output at its predicted location. 

\myparagraph{Spatial context prediction module.}
The spatial context prediction module predicts cell-specific K-function vectors (as defined in Eq.~\ref{eq:cell-k-func}
). The intuition is that by predicting spatial context, the learnt feature representation is spatial-context-aware and can help detection and classification modules. This will be verified in the experimental section via an ablation study.

We predict an 18 dimensional K-function vector for each cell. For ground truth, we use the ground truth cell detection mask generated for the cell detection task. For each positive pixel, i.e., the pixels within the dilated components, we compute the K-function vector.
Our spatial-context-prediction block output has the same spatial resolution as the input image, and has 18 channels corresponding to the 18 dimensional K-function vectors. No additional activation is needed. Only the predictions at the positive pixels are compared with the ground truth K-function vectors. 

To compare two K-function vectors, we can use the Kolmogorov–Smirnov test \cite{kstest} for comparing sample estimates of cumulative distribution functions. Formally, it is the supreme norm of the difference between the prediction and ground truth:
$$KM(K_{pred}, K_{gt}) = \sup\nolimits_r |K_{pred}(r) - K_{gt}(r)|.$$
In practice, we found out that the supreme norm is less efficient and we use L1 norm as a surrogate loss.


\myparagraph{Deep clustering module.}
The spatial context prediction module learns representations of spatial context. However, in practice, we observe that it does not collaborate well with the detection and classification modules. We hypothesize that the appearance feature and the spatial context feature do not fuse well. This could be due to the very different natures of different tasks. The detection and classification tasks are per-pixel classification tasks with a small number of classes. Whereas the spatial context prediction task is a regression problem of high dimensional output. 

In order to re-calibrate the appearance features and spatial features learnt from these different tasks, we propose a deep clustering module, which plays the role of self-supervision and lies in the mid-ground between the other modules. The deep clustering module executes a per-pixel classification task, but with a higher number of pseudo-classes derived from both spatial and appearance information.
Deep clustering is known to enhance feature representation \cite{caron:2018:eccv:deep-cluster,Chang:2020:CVPR:deep-clustering:weak-supervised} especially for unsupervised and weakly-supervised tasks. This also fits our setting; we do not have a golden standard of sub-categories of cells. Instead, we derive the sub-classes dynamically based on the feature representation and train the model to predict them.

In particular, we apply k-means clustering on the intermediate feature representation to obtain pseudo-sub-classes that further divide cells of each class. Each class of cells is stratified into 5 pseudo-sub-classes. The deep-clustering block then learns to predict these pseudo labels. The design of this block 
is similar to the cell classification module.
We note that the intermediate feature representation for clustering is a concatenation of the features from the deep-clustering block and the spatial-context-prediction block. This ensures that the clustering, and thus the learnt feature representation, integrates both spatial and appearance information in a well-calibrated way.


The clustering and derived pseudo-sub-class labels are re-generated on the start of every epoch. 
When running k-means clustering, the clusters' centroids are initialized with the previous epoch's centroids to avoid excess jumping between cluster mapping. 
The training is very similar to the cell classification task, using DICE loss. 

\myparagraph{The overall loss of our model} is the weighted sum of losses from all four modules: $$\mathcal{L}=\lambda_1\mathcal{L}_{\text{Det}}+\lambda_2\mathcal{L}_{\text{Class}}+\lambda_3\mathcal{L}_{\text{Spatial}}+\lambda_4\mathcal{L}_{\text{DeepCluster}}.$$ 
In practice, we simply set all weights to one. 

\myparagraph{Technical details.}
We finalize this subsection with a few more details of our model architecture. The feature extractor has VGG-16 encoder and has 4 decoder blocks. The last decoder block is a single deconvolution layer with 96 output channels. The detection, classification, deep clustering, and spatial context blocks, all have a similar architecture: two $3\times3$ convolutions with $64$ channels output, followed by $1\times1$ convolution to give the final output. Each convolution except for last is followed by a ReLU activation. More details are in Section~\ref{sec:experiments} and supplementary material.
\section{Experiments}
\label{sec:experiments}
\setlength{\tabcolsep}{4pt}
\begin{table*}
\small
\begin{center}
\caption{Results on all three datasets. For each dataset, we report five scores: F-scores for individual classes (inflammatory, epithelial, stromal), mean F-score over all three classes (Mean), and detection F-score over all cells (Det.). For each score, the best method is highlighted with bold fonts.}
\label{table:all-results}
\begin{tabular}{|p{0.140\linewidth}|p{0.04\linewidth}|p{0.04\linewidth}|p{0.04\linewidth}|p{0.04\linewidth}|p{0.04\linewidth}|p{0.04\linewidth}|p{0.04\linewidth}|p{0.04\linewidth}|p{0.04\linewidth}|p{0.04\linewidth}|p{0.04\linewidth}|p{0.04\linewidth}|p{0.04\linewidth}|p{0.04\linewidth}|p{0.04\linewidth}|}
\hline
& \multicolumn{5}{p{0.26\linewidth}|}{\centering {BRCA-M2C}}  & \multicolumn{5}{p{0.26\linewidth}|}{\centering {Consep}} & \multicolumn{5}{p{0.26\linewidth}|}{\centering {SEER-Lung}}\\
\hline
 Method  & \centering{Infl.} & \centering{Epi.} & \centering{Stro.} & \centering{Mean} & \centering{Det.}  & \centering{Infl.} & \centering{Epi.} & \centering{Stro.} & \centering{Mean} & \centering{Det.} & \centering{Infl.} & \centering{Epi.} & \centering{Stro.} & \centering{Mean}& \centering{Det.}\cr
\hline
U-Net 
& \centering{0.498} & \centering{0.744} & \centering{0.476} & \centering{0.572} & \centering{0.838}
& \centering{0.681} & \centering{0.613} & \centering{0.561} & \centering{0.618} & \centering{0.724}
& \centering{0.779} & \centering{0.809} & \centering{0.571} & \centering{0.720} & \centering{0.856} 
\cr
Faster-RCNN \cite{faster-rcnn:2018:NIPS}  
& \centering{0.572} & \centering{0.718} & \centering{0.490} & \centering{0.594} & \centering{0.806}
& \centering{0.259} & \centering{0.523} & \centering{0.446} & \centering{0.410} & \centering{0.492} 
& \centering{0.192} & \centering{0.769} & \centering{0.411} & \centering{0.457} & \centering{0.616} 
\cr
Cascade RCNN \cite{cai:cascade-rcnn:2018:CVPR}  
& \centering{0.564} & \centering{0.708} & \centering{0.505} & \centering{0.592} & \centering{0.796}
& \centering{0.644} & \centering{0.633} & \centering{0.515} & \centering{0.597} & \centering{0.682}
& \centering{0.710} & \centering{0.793} & \centering{0.454} & \centering{0.653} & \centering{0.759}
\cr
PointSeg\cite{Qu:midl:2019:point_seg}
& \centering{0.249} & \centering{0.407} & \centering{0.300} & \centering{0.319} & \centering{0.538}
& \centering{0.104} & \centering{0.603} & \centering{0.210} & \centering{0.306} & \centering{0.435} 
& \centering{0.748} & \centering{0.773} & \centering{0.537} & \centering{0.686} & \centering{0.848} 
\cr
HoverNet-Weakly 
& \centering{0.582} & \centering{0.702} & \centering{0.513} & \centering{0.599} & \centering{0.817}
& \centering{0.549} & \centering{0.377} & \centering{0.365} & \centering{0.431} & \centering{0.518} 
& \centering{\textbf{0.823}} & \centering{0.808} & \centering{0.535} & \centering{0.722} & \centering{0.846} 
\cr
HoverNet \cite{graham:2019:hover}  
& \centering{-} & \centering{-} & \centering{-} & \centering{-} & \centering{-}
& \centering{0.633} & \centering{0.651} & \centering{0.624} & \centering{0.636} & \centering{0.730}
& \centering{-} & \centering{-} & \centering{-} & \centering{-} & \centering{-}
\cr
\hline
MCSpatNet  
& \centering{\textbf{0.635}} & \centering{\textbf{0.785}} & \centering{\textbf{0.553}} & \centering{\textbf{0.658}} & \centering{\textbf{0.849}}
& \centering{\textbf{0.724}} & \centering{\textbf{0.695}} & \centering{\textbf{0.628}} & \centering{\textbf{0.682}} & \centering{\textbf{0.762}} 
& \centering{0.799} & \centering{\textbf{0.817}} & \centering{\textbf{0.600}} & \centering{\textbf{0.739}} & \centering{\textbf{0.860}}
\cr
\hline
\end{tabular}
\end{center}
\end{table*}

\setlength{\tabcolsep}{4pt}
\begin{table*}[t!]
\small
\begin{center}
\caption{Ablation study on the 3 datasets. The metrics and the highlight convention are the same as in Table \ref{table:all-results}. }
\label{table:all-ablation}
\begin{tabular}{|p{0.14\linewidth}|p{0.04\linewidth}|p{0.04\linewidth}|p{0.04\linewidth}|p{0.04\linewidth}|p{0.04\linewidth}|p{0.04\linewidth}|p{0.04\linewidth}|p{0.04\linewidth}|p{0.04\linewidth}|p{0.04\linewidth}|p{0.04\linewidth}|p{0.04\linewidth}|p{0.04\linewidth}|p{0.04\linewidth}|p{0.04\linewidth}|}
\hline
& \multicolumn{5}{p{0.26\linewidth}|}{\centering {BRCA-M2C}}  & \multicolumn{5}{p{0.26\linewidth}|}{\centering {Consep}} & \multicolumn{5}{p{0.26\linewidth}|}{\centering {SEER-Lung}}\\
\hline
 Method  & \centering{Infl.} & \centering{Epi.} & \centering{Stro.} & \centering{Mean} & \centering{Det.}  & \centering{Infl.} & \centering{Epi.} & \centering{Stro.} & \centering{Mean} & \centering{Det.} & \centering{Infl.} & \centering{Epi.} & \centering{Stro.} & \centering{Mean}& \centering{Det.}\cr
\hline
\rowcolor{LightGray}
U-Net 
& \centering{0.498} & \centering{0.744} & \centering{0.476} & \centering{0.572} & \centering{0.838}
& \centering{0.681} & \centering{0.613} & \centering{0.561} & \centering{0.618} & \centering{0.724} 
& \centering{0.779} & \centering{0.809} & \centering{0.571} & \centering{0.720} & \centering{0.856} 
\cr
U-Net+Deep Clus.
& \centering{0.593} & \centering{0.763} & \centering{0.505} & \centering{0.620} & \centering{0.851}
& \centering{0.613} & \centering{0.644} & \centering{0.504} & \centering{0.587}  & \centering{0.714} 
& \centering{0.768} & \centering{0.804} & \centering{0.576}  & \centering{0.716} & \centering{0.849} 
\cr
\rowcolor{LightGray}
U-Net+Spat.~Pred.
& \centering{0.597} & \centering{0.771} & \centering{0.507} & \centering{0.625}& \centering{\textbf{0.853}} 
& \centering{0.682} & \centering{0.650} & \centering{0.522} & \centering{0.618} & \centering{0.761} 
& \centering{0.787} & \centering{0.809} & \centering{0.587} & \centering{0.728} & \centering{0.851} 
\cr
\hline
Using NN Dist.
& \centering{\textbf{0.641}} & \centering{0.698} & \centering{0.447} & \centering{0.595} & \centering{0.829} 
& \centering{0.656} & \centering{0.616} & \centering{0.568} & \centering{0.613} & \centering{0.719} 
& \centering{0.752} & \centering{0.808} & \centering{0.558} & \centering{0.706} & \centering{0.848} 
\cr
\rowcolor{LightGray}
Using Density
& \centering{0.563} & \centering{0.753} & \centering{0.525} & \centering{0.614} & \centering{0.851} 
& \centering{0.686} & \centering{0.687} & \centering{0.624} & \centering{0.666} & \centering{\textbf{0.764}} 
& \centering{0.797} & \centering{0.807} & \centering{0.572} & \centering{0.725} & \centering{0.850} 
\cr
\hline
MCSpatNet
& \centering{0.635} & \centering{\textbf{0.785}} & \centering{\textbf{0.553}} & \centering{\textbf{0.658}} & \centering{0.849}
& \centering{\textbf{0.724}} & \centering{\textbf{0.695}} & \centering{\textbf{0.628}} & \centering{\textbf{0.682}} & \centering{0.762}
& \centering{\textbf{0.799}} & \centering{\textbf{0.817}} & \centering{\textbf{0.600}} & \centering{\textbf{0.739}} & \centering{\textbf{0.860}} 
\cr
\hline
\end{tabular}
\end{center}
\end{table*}
We evaluate our method, MCSpatNet, on three datasets of different cancer types: breast cancer, lung cancer, and colorectal cancer. The breast cancer dataset, BRCA-M2C, consists of 120 patches belonging to 113 patients, collected from TCGA \cite{tcga}. The lung cancer dataset, SEER-Lung, is a collection of 57 patches from the SEER cohort \cite{seer-ref}. Each patch is sampled from different whole slide images or tissue samples to maximize the generalizability. The colorectal cancer dataset, Consep \cite{graham:2019:hover}, is publicly available. It has 41 patches. The patches from all 3 datasets are of size $\approx 500\times500$ at $20$x magnification; they are large enough to provide spatial context. The lung and breast cancer datasets are annotated by pathologists with ground truth points at approximate centers of cells with an associated class: inflammatory, epithelial, or stromal. It is worth mentioning that all the epithelial cells in these patches are tumor cells.
The Consep dataset additionally has nuclei contour masks. Details of these datasets are in the supplemental material. 


\myparagraph{Implementation details of MCSpatNet.}
We train the model on patches taken at 20x magnification, which is around 0.5 microns per pixel. We use dilated ground truth dot masks with dilation up to 9 pixels 
In all datasets, the model is trained on the 3 major classes: inflammatory, epithelial, and stromal cells. For the deep clustering, we use kmeans. Each cell class has k=5 clusters (15 in total). 

To train the spatial context module, we use an R-package to generate the ground truth K-functions with border correction. These K-functions only need to be computed once before training starts. It takes around 0.1 min per patch. The radius $r$'s range=$[0,90]$ with step=$15$ pixels. To get the formula as in Eq.~\ref{eq:k-func}, the returned K-function is multiplied by $\frac{\text{number of cells in region}}{area\times n_{max}}$, where $n_{max}$ is a constant set to $100$.  
During inference, we apply a threshold=$0.5$ on the detection output and get rid of tiny connected components with area less than $5$ pixels. The location of the predicted cells are then the centroids of the resulting connected components. 


\myparagraph{Evaluation.}
We train and evaluate our method against state-of-the-arts on all three datasets. We split each training dataset into train and validation sets and evaluate all methods on the same split. 
The performance is evaluated with the F-score metric. We report the F-scores on the detection and classification tasks.
Similar to existing approaches \cite{graham:2019:hover,sirinukunwattana:itmi:2016:point_class}, we say a predicted cell is true positive (TP) if it is within 6 pixels (around 3 microns) from one of the ground truth points. Otherwise it is a false positive (FP). A ground truth point is false negative (FN) if it does not have a nearby prediction. F-score is then calculated as $\frac{TP}{TP + 0.5(FP+FN)}$.

The detection F-score is computed over \textit{all} detected cells regardless of their class.
The classification F-score 
is evaluated on cells of each class (inflammatory F-score, epithelial F-score and stromal F-score). We also report the mean F-score over the three classes as an overall metric for the classification performance. 

\myparagraph{Baselines.} We compare with several SOTA methods which can jointly segment/detect and classify cells.
\textbf{U-Net} is a baseline method for joint detection and classification. It uses a U-Net architecture with VGG-16 backbone. It is essentially our method (MCSpatNet) without the spatial prediction and deep clustering modules. We also compare against SOTA computer vision multi-class detection algorithms, e.g.,~\textbf{Faster-RCNN} \cite{faster-rcnn:2018:NIPS} and \textbf{Cascade RCNN} \cite{cai:cascade-rcnn:2018:CVPR}. 

Aside from detection based methods, segmentation-based methods can also be applied. Since the ground truth nuclei masks are not available, we apply a SOTA weakly supervised nuclei segmentation method (\textbf{PointSeg}) \cite{Qu:midl:2019:point_seg}, which only requires point annotation for training. 
To classify the segmentation results, we train a CNN classifier (\textbf{SSPP}) \cite{sirinukunwattana:itmi:2016:point_class} on local patches enclosing each predicted cell segment. 
We also apply 
\textbf{HoverNet} \cite{graham:2019:hover}, a SOTA joint segmentation and classification method. Training HoverNet requires fully annotated nucleus masks, which is only available for Consep dataset. We only apply the full HoverNet on Consep. Meanwhile, for all three datasets, we apply a weakly-supervised HoverNet (\textbf{HoverNet-Weakly}) by training on pseudo-masks acquired from point-annotation-anchored superpixels. More details can be found in the supplemental material. 

\begin{figure*}[t!]
    \begin{center}
       \includegraphics[width=1\linewidth]{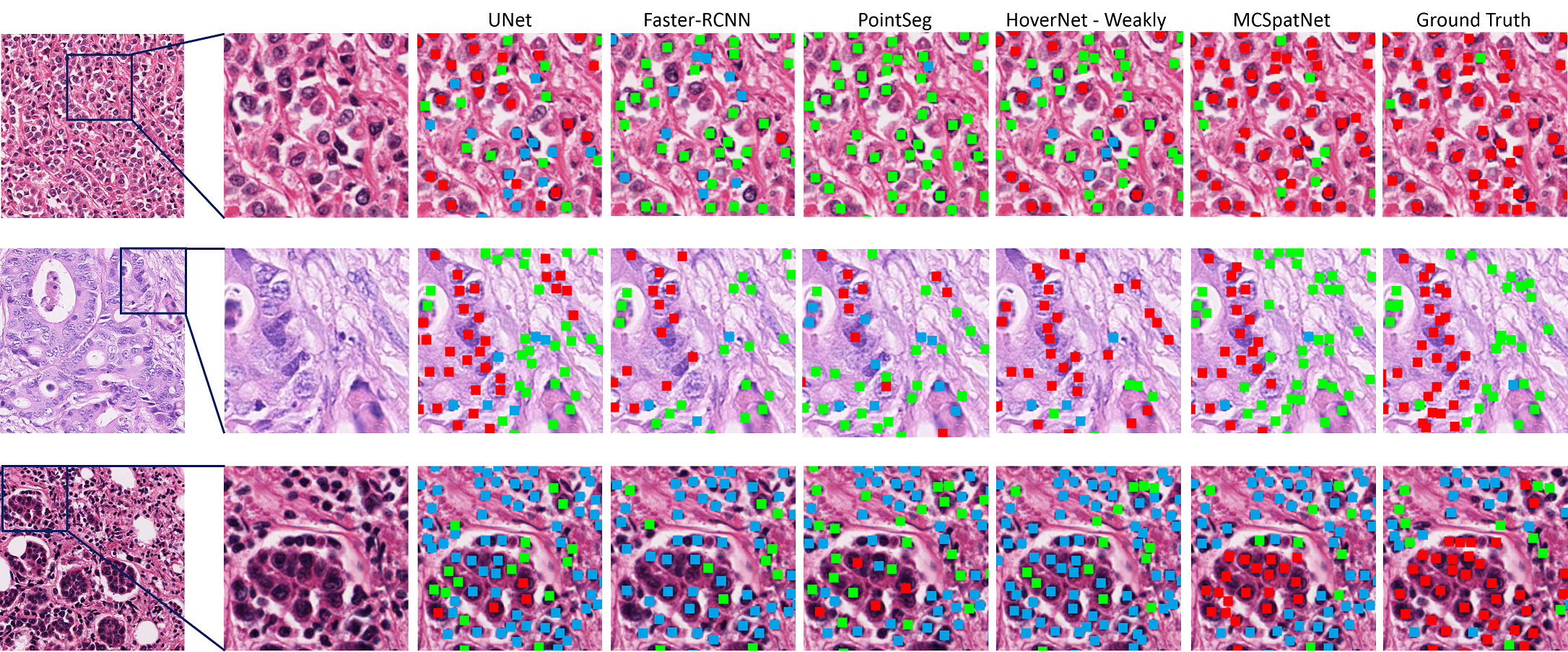}
    \end{center}
   \caption{Qualitative Results. Blue, red, and  green dots represent  inflammatory,  epithelial,  and  stromal cells, respectively.}
\label{fig:teaser4}
\end{figure*}
\myparagraph{Results and discussions.}
Table \ref{table:all-results} shows the results of our method and all baselines on all three datasets. Our method consistently outperform all baselines on different metrics. Generally, we observe a bigger advantage in the classification task (inflammatory, epithelial, stromal F-scores and mean F-score) than in the detection task (Det. F-score). This is expected as spatial context is used by pathologists to help determine cell classes in ambiguous situation, whereas appearance is the main cue for detection.  

Figure \ref{fig:teaser4} shows the qualitative results of our method and different baselines. 
In all patches, the appearance of cells are not very discriminating and baseline models are unable to correctly classify them. Our model is able to differentiate epithelial vs.~stromal cells and epithelial vs.~inflammable cells. More results are in the supplemental material.

\myparagraph{Ablation study.}
We evaluate our proposed method against a few variations to show the efficacy of different components. In all the runs we use the same experimental setting as above. 
We first evaluate the efficacy of the proposed two novel modules: spatial prediction and deep clustering. To this end, we evaluate 3 baselines: U-Net (the model with both modules removed), U-Net plus deep clustering module, and U-Net plus spatial prediction module. See Table \ref{table:all-ablation} for the results. Comparing U-Net with the second and third baselines, we observe both modules contribute to performance improvement. Furthermore, the full version of our model, MCSpatNet, outperforms all three baselines. This establishes the necessity of both modules.

To further investigate the efficacy of the proposed K-function, we compare with using different spatial context descriptors: nearest neighbor distance function and density function.
For the first baseline, in the spatial prediction module, we replace the K-function vector with the distance to the nearest neighbor cell of each of the 3 classes. For the second baseline, we replace the K-function vector with the density of the cells from each of the 3 classes within the neighborhood area. 
As shown in Table \ref{table:all-ablation}, K-function based spatial prediction outperforms both of these baselines. This demonstrates that the K-function provides richer spatial information and helps the model learn a spatial-context-aware representation that best suits the cell detection and classification tasks.

More ablation studies are in the supplemental material.

\myparagraph{Statistical significance.}
To verify that the benefit of our method is robust, we ran our method and two top baselines: UNet and HoverNet-Weakly, for three more splits on BRCA-M2C (Table \ref{table:statistical}). 
Our method is consistently the best over all scores. 
Furthermore, we ran a paired T-test comparing our method with these baselines. We highlight a baseline result if it is not statistically significantly different from ours (i.e., p-value $>0.05$). 
These ``close second'' results often have much lower average, but  high standard deviation; they are very unstable.

\begin{table}[t!]
\small
 \caption{Average and standard deviation of scores on four random splits of BRCA-M2C.}
 \label{table:statistical}
\centering
\setlength\tabcolsep{5pt}
\begin{tabular}{|p{0.09\linewidth}|p{0.2\linewidth}|p{0.2\linewidth}|p{0.19\linewidth}|}
\hline
\centering{} & \centering{MCSpatNet} & \centering{U-Net} & \centering{Hovernet-Weakly} \cr
\hline
\rowcolor{LightGray}
\centering{Infl.} & \centering{$\mathbf{0.71 \pm 0.07}$} & \centering{$\mathbf{0.58 \pm 0.22}$} & \centering{$0.54 \pm 0.04$}
\cr
\hline
\centering{Epi.} & \centering{$\mathbf{0.76 \pm 0.03}$} & \centering{$0.72 \pm 0.02$} & \centering{$\mathbf{0.55 \pm 0.19}$} 
\cr
\hline
\rowcolor{LightGray}
\centering{Stro.} & \centering{$\mathbf{0.56 \pm 0.08}$} & \centering{$0.51 \pm 0.09$} & \centering{$0.49 \pm 0.08$} 
\cr
\hline
\centering{Mean} & \centering{$\mathbf{0.67 \pm 0.04}$} & \centering{$\mathbf{0.60 \pm 0.10}$} & \centering{$0.53 \pm 0.06$}
\cr
\hline
\rowcolor{LightGray}
\centering{Det.} & \centering{$\mathbf{0.85 \pm 0.02}$} & \centering{$0.84 \pm 0.01$} & \centering{$0.78 \pm 0.03$}
\cr
\hline
\end{tabular}
\vspace{-.15in}
 \end{table}

\section{Conclusion}
In this paper, we propose a novel method for joint cell detection and classification. The novel contribution is to explicitly introduce spatial context information and train the model to learn a spatial-context-aware cell representation of cells.
We also use a deep clustering method to better re-calibrate the spatial and appearance features. The proposed method outperforms different SOTA methods, demonstrating the significance of spatial context.

In the future, we will extend the method to other forms of contextual information, e.g., topology \cite{aukerman2020persistent,abousamra2021localization}, and other stainings, e.g., multiplex immunohistochemistry \cite{abousamra2020weakly,fassler2020deep}.

\myparagraph{Acknowledgement.}
We thank anonymous reviewers for their constructive feedback.
This work was partially supported by grants NSF IIS-1909038, CCF-1855760, NCI 1R01CA253368-01, 
NCI UH3-CA225021 and NCI U24CA215109  
as well as generous donor  funding from Betsy and Bob Barton.
 
{\small
\bibliographystyle{ieee_fullname}
\bibliography{references-abbrev,egbib,chao}
}

\appendix


\section{Model Architecture}
\label{sec:supp:architecture}
\begin{figure*}[t]
    \begin{center}
      \includegraphics[width=1\linewidth]{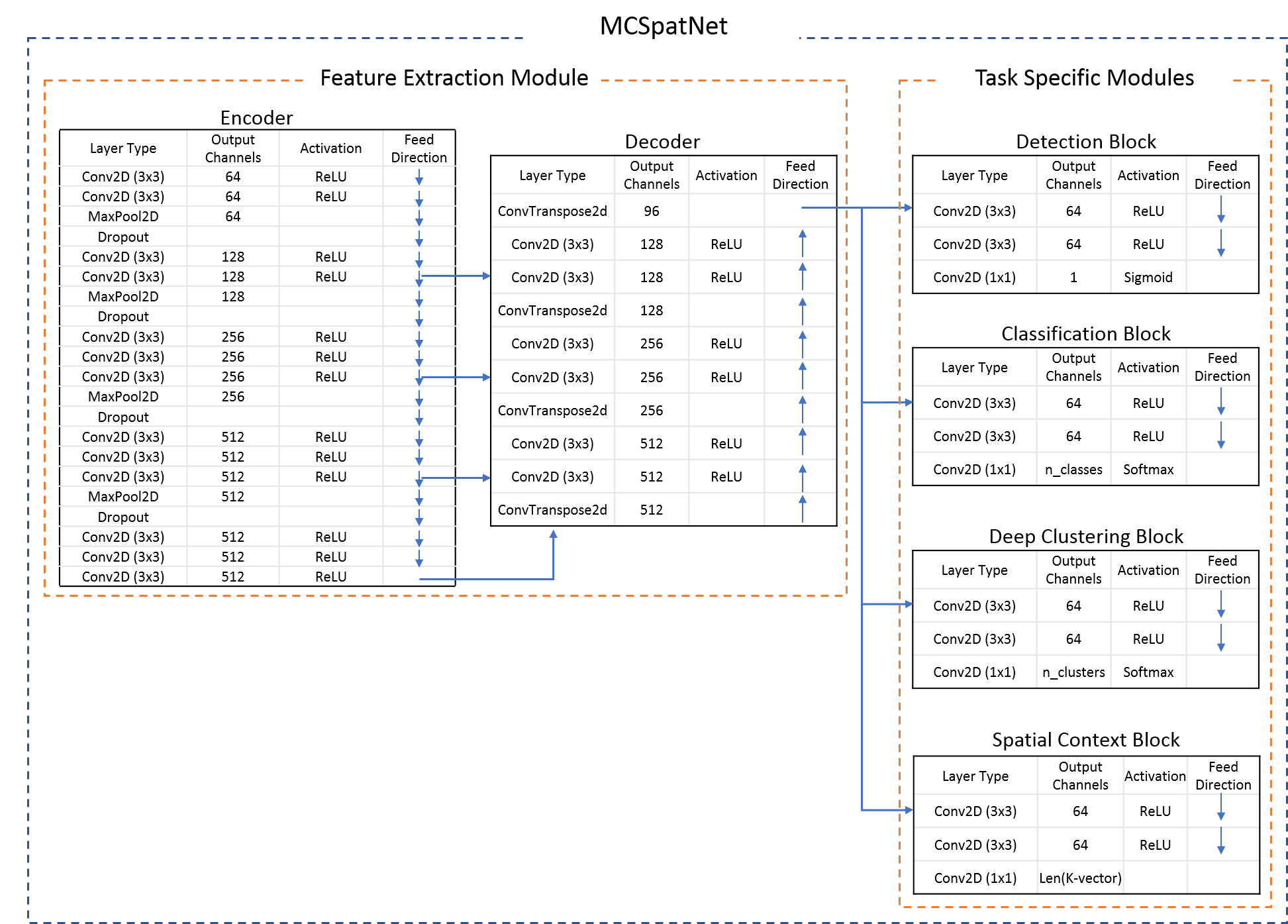}
  \end{center}
  \caption{Model architecture details. }
\label{fig:arch-table}
\end{figure*}

Fig.~\ref{fig:arch-table} shows details of the proposed deep neural network architecture.

\section{Datasets}
\label{sec:datasets}
\setlength{\tabcolsep}{4pt}
\begin{table*}[t]
\small
\begin{center}
\caption{Datasets statistics. For each dataset, we report number of training/validation/test patches. For each class, we also report numbers of cells in training/validation/test sets.}
\label{table:datasets}
\begin{tabular}{|p{0.22\linewidth}|p{0.16\linewidth}|p{0.16\linewidth}|p{0.16\linewidth}|p{0.16\linewidth}|}
\hline
 Dataset  & \centering{N Patches}  & \centering{Inflam.} & \centering{Epi.} & \centering{Stroma}  \cr
\hline
\rowcolor{LightGray}
BRCA-M2C
& \centering{80 / 10 / 30} 
& \centering{3541 / 1358 / \; 960} 
& \centering{9956 / \; 733 / 6109} 
& \centering{5150 / 1042 / 1789} 
\cr
SEER-Lung 
& \centering{37 /  5 / 15} 
& \centering{5499 / \; 726 / 3085} 
& \centering{10875 / 1601 / 4906} 
& \centering{5708 / \; 407 / 1871}  
\cr
\rowcolor{LightGray}
Consep 
& \centering{22 /  5 / 14} 
& \centering{2982 / \; 652\;/ 1537} 
& \centering{\; 4987 / \; 376 / 3113} 
& \centering{4394 / 1343 / 3640} 
\cr

\hline
\end{tabular}
\end{center}
\end{table*}

\begin{figure*}[t]
    \begin{center}
       \includegraphics[width=1\linewidth]{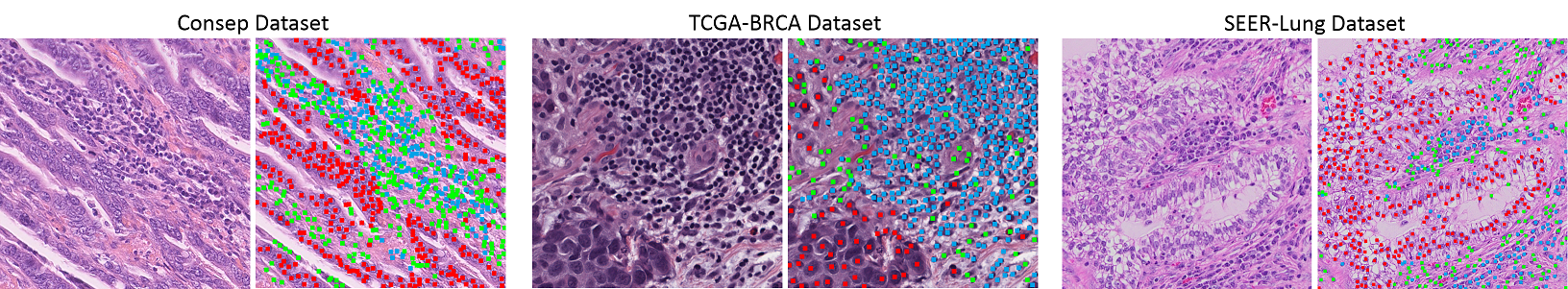}
   \end{center}
   \caption{Sample patches and the corresponding point annotation. Colors indicate cell types: red=epithelial cells, blue=inflammatory cells, green=stromal cells.}
\label{fig:supp-datasets-samples}
\end{figure*}

We provide more statistics of the datasets in Table \ref{table:datasets}. In Figure \ref{fig:supp-datasets-samples}, we provide sample patches and annotations from the three datasets. It should be noted that even point annotation is expensive. It takes around 30 minutes to annotate a patch by our pathologist annotator.

\section{Baselines}
\label{sec:baselines}
We provide details of the baseline algorithms.

\myparagraph{HoverNet-Weakly.}
Our weakly-supervised HoverNet baseline is trained on pseudo nuclei masks generated using superpixels. We apply the SLIC \cite{achanta:2012:itmi:slic} algorithm to partition the input image into superpixels. Each superpixel that contains an annotation point is considered the mask of a nucleus. In order to account for the variation in cell size and shape across different classes, we fine-tune the superpixel algorithm parameters in a class-specific manner.

\begin{figure*}[t]
    \begin{center}
      \includegraphics[width=0.8\linewidth]{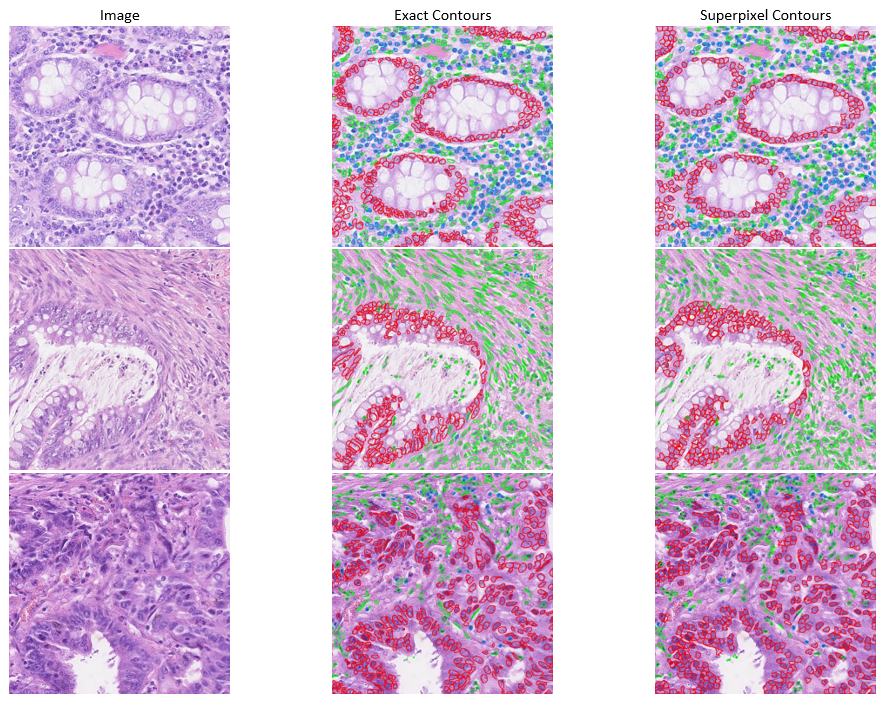}
  \end{center}
  \caption{Sample images from Consep dataset, the exact segmentation masks (exact contours) provided by the original dataset, and superpixel-based pseudo masks (superpixel contours). }
\label{fig:superpixel_consep}
\end{figure*}

In Figure \ref{fig:superpixel_consep}, we visualize the superpixels for different classes of the Consep dataset. For reference, we also show the ground truth segmentation mask which was provided in Consep dataset. We observe that the superpixel-based pseudo-masks reasonably capture the scale and shape of the cells, but lose details along the boundary.

\myparagraph{Faster-RCNN.}
For the baseline Faster-RCNN,
we use the standard C4 architecture \cite{faster-rcnn:2018:NIPS} with a ResNet101 feature encoder. All models were trained for 50,000 iterations. The only significant modification was in the maximal number of predicted instances, which we set to 1000 to allow for patches with a large number of cells, and the test ROI confidence threshold, which we set to 0.7 during inference. 

\myparagraph{Cascade-RCNN}
For the baseline Cascade-RCNN, we use the same encoder as in the paper \cite{cai:cascade-rcnn:2018:CVPR}, which is a Resnet101 based feature pyramid network. All models were trained for 50,000 iterations. To tune the model for this data domain, we set the anchor sizes to [8, 16, 32, 64, 128]. We find that the iterative process of detecting the bounding boxes used by Cascade-RCNN results in too many correct predictions failing the confidence threshold. As such, we both increase the maximum number of detections per image to 10000 and decrease the test ROI confidence threshold to 0.6.   

\myparagraph{PointSeg.}
The PointSeg baseline is based on the weakly-supervised  segmentation model from \cite{Qu:midl:2019:point_seg}. We use this model to segment nuclei from an input patch. Next, we use the SSPP classification network \cite{sirinukunwattana:itmi:2016:point_class} to perform classification. As in the original paper this network is trained on 27x27 20x magnification patches based on the point annotations. To account for variance in detected center, each point has up to 9 training patches selected, one centered at the point, and the other eight with centers up to 6 pixels away from the annotation point in varying directions. At inference stage, we use the weakly supervised segmentation network to predict segments of nuclei. For each segment, we use its centroid as the predicted cell location, extract a patch centered at the location, and apply the  classification network to predict its class.

\section{Average K-function Curves}
\label{sec:k-curves}
\begin{figure*}[t]
\begin{center}
  \includegraphics[width=1\linewidth]{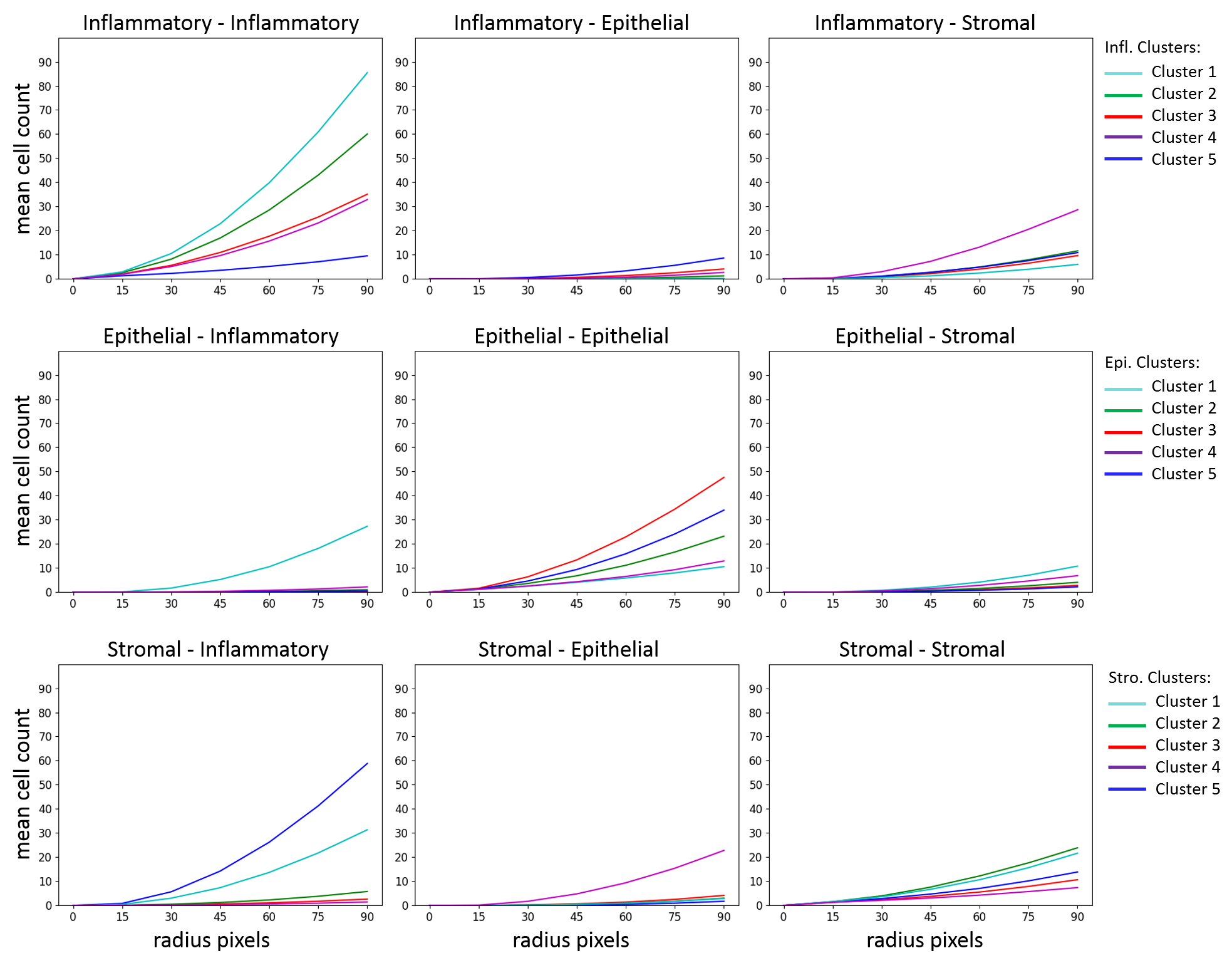}
\end{center}
  \caption{Average K-curves per cell type pair in each cluster. Each plot is for a pair of cell classes. The colored curves represent the K-curve for that pair in the different clusters. This clearly shows that different clusters demonstrate different cellular spatial behavior.}
\label{fig:k-curve-avg-cluster}
\end{figure*}

In Fig.~\ref{fig:k-curve-avg-cluster}, the average pairwise K-curves are plotted for each pair of classes in each cluster. It shows how different clusters demonstrate different spatial behavior across pairs of cell classes. This also shows that the K-function across clusters is able to capture the variation in the spatial context.

\section{Additional Experiments}
\label{sec:experiments}
We provide additional ablation studies on the hyperparameters selection for the maximum radius of the K-function and the number of clusters in the deep clustering module.

\paragraph{Varying K-function radius.}
Table~\ref{table:radius-variation} shows the results of varying the maximum radius of the K-function on the BRCA-M2C dataset. We experiment with radii of $60$, $90$, and $120$ pixels. There is practically no difference in the detection F-score but there is a clear difference in the classification F-scores, which again shows the advantage of the spatial context learning to the classification task. $60$ and $120$ appear to be too small and too large radii. The best performance is at $90$ pixels which is the setting we used in our experiments.

\paragraph{Varying number of clusters.}
Table~\ref{table:cluster-variation} shows the results of varying the number of clusters in the deep clustering module. We experiment on the BRCA-M2C dataset with $3$, $5$, and $7$ clusters. Again, the variation has no effect on the detection task. There are slight differences on the classification results with the best being at 5 clusters, which is the setting used in our experiments.

\setlength{\tabcolsep}{4pt}
\begin{table}[hb!]
\small
\begin{center}
\caption{Ablation study: Varying the K-function maximum radius on the BRCA-M2C dataset. Evaluation with F-scores per class, mean F-score over classes and detection F-score over all cells. Infl.: Inflammatory cells, Epi.: Epithelial cells, Stro.: Stromal cells, Mean: the mean F-score over the 3 classes, Det.: detection F-score.}
\label{table:radius-variation}
\begin{tabular}{|p{0.15\linewidth}|p{0.09\linewidth}|p{0.09\linewidth}|p{0.09\linewidth}|p{0.09\linewidth}|p{0.09\linewidth}|}
\hline
Radius & \centering{Infl.} & \centering{Epi.} & \centering{Stro.} & \centering{Mean} & \centering{Det.}
\cr
\hline
60 & \centering{0.612} & \centering{0.777} & \centering{0.526} & \centering{0.638} & \centering{0.851}
\cr
\hline
90 & \centering{0.635} & \centering{0.785} & \centering{0.553} & \centering{0.658} & \centering{0.849}
\cr
\hline
120 & \centering{0.594} & \centering{0.761} & \centering{0.507} & \centering{0.621} & \centering{0.854}
\cr
\hline
\end{tabular}
\end{center}
\end{table}

\setlength{\tabcolsep}{4pt}
\begin{table}[t!]
\small
\begin{center}
\caption{Ablation study: Varying the number of clusters in the deep clustering module on the BRCA-M2C dataset. Evaluation with F-scores per class, mean F-score over classes and detection F-score over all cells. Infl.: Inflammatory cells, Epi.: Epithelial cells, Stro.: Stromal cells, Mean: the mean F-score over the 3 classes, Det.: detection F-score.}
\label{table:cluster-variation}
\begin{tabular}{|p{0.15\linewidth}|p{0.09\linewidth}|p{0.09\linewidth}|p{0.09\linewidth}|p{0.09\linewidth}|p{0.09\linewidth}|}
\hline
Clusters & \centering{Infl.} & \centering{Epi.} & \centering{Stro.} & \centering{Mean} & \centering{Det.}
\cr
\hline
3 & \centering{0.622} & \centering{0.760} & \centering{0.508} & \centering{0.630} & \centering{0.850}
\cr
\hline
5 & \centering{0.635} & \centering{0.785} & \centering{0.553} & \centering{0.658} & \centering{0.849}
\cr
\hline
7 & \centering{0.630} & \centering{0.776} & \centering{0.526} & \centering{0.644} & \centering{0.850}
\cr
\hline
\end{tabular}
\end{center}
\end{table}

\section{Additional Qualitative Results}
\label{sec:qualitative}
\begin{figure*}[t]
       \includegraphics[width=1\linewidth]{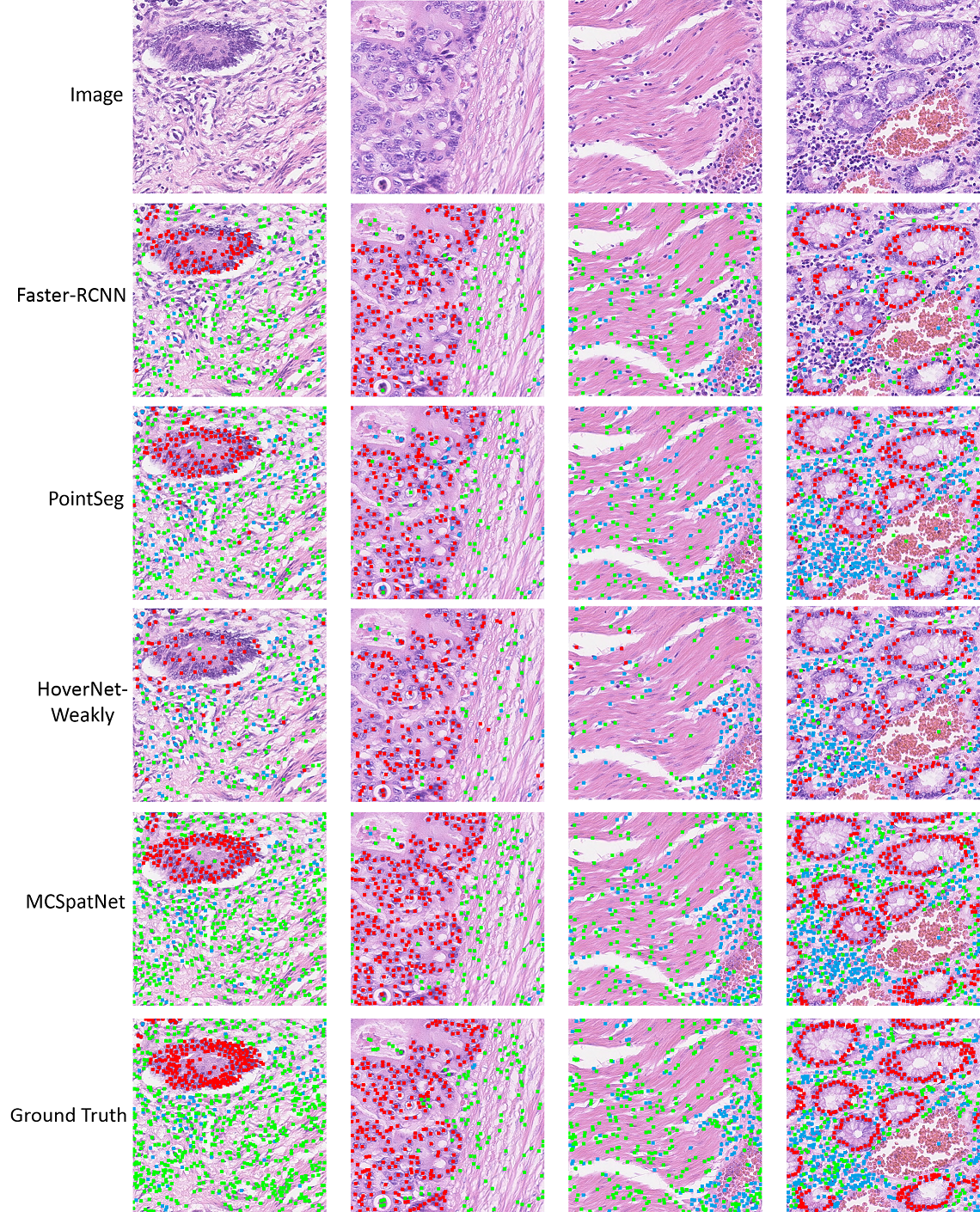}
   \caption{Qualitative results comparing against baseline methods in dense tumor regions. Top row are the original image patches. Bottom row are the ground truths. (Blue=inflammatory cells, Red=epithelial cells, Green=stromal cells.)}
\label{fig:supp-qualitative-compare1}
\end{figure*}

\begin{figure*}[t]
       \includegraphics[width=1\linewidth]{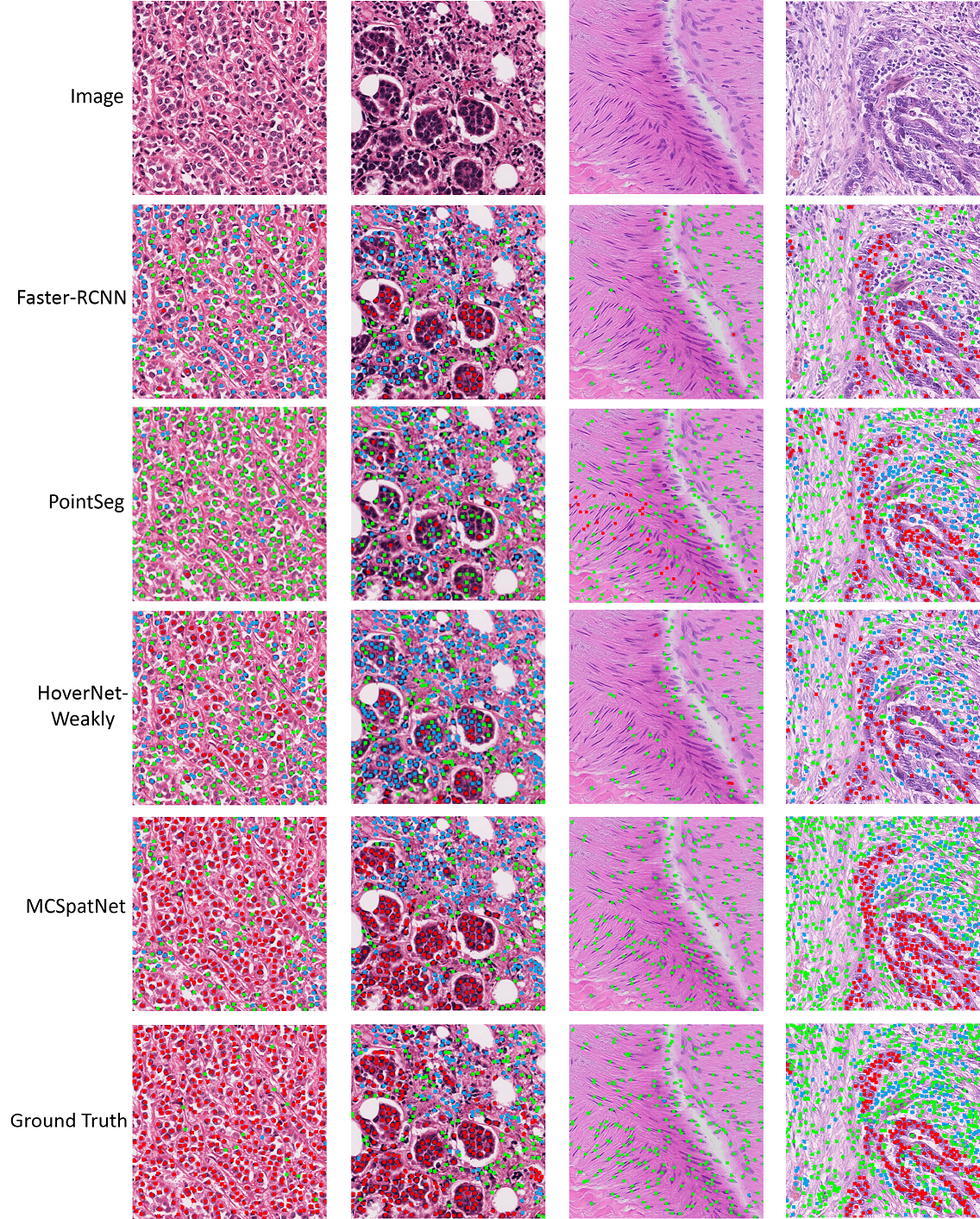}
   \caption{Qualitative results comparing against baseline methods in various tissue regions. Top row are the original image patches. Bottom row are the ground truths. (Blue=inflammatory cells, Red=epithelial cells, Green=stromal cells.)}
\label{fig:supp-qualitative-compare2}
\end{figure*}

We provide more qualitative examples to demonstrate the benefit of our proposed method, MCSpatNet, in Figures~\ref{fig:supp-qualitative-compare1} and \ref{fig:supp-qualitative-compare2}. Both figures show that MCSpatNet has better classification accuracy. Figure~\ref{fig:supp-qualitative-compare1} demonstrates how MCSpatNet can better detect and classify cells even in very dense tumor nest regions. Figure~\ref{fig:supp-qualitative-compare2} shows that MCSpatNet performs equally well in various tissue regions.





\end{document}